\documentclass[12pt]{article}
\usepackage{geometry}
\usepackage{hyperref}
\usepackage{amsmath}
\usepackage{graphicx}
\usepackage[numbers]{natbib}
\usepackage{enumitem}
\usepackage{booktabs}

\newcommand{\datasetlink}{\href{https://huggingface.co/datasets/biographs/jokes_on_us}{Hugging~Face}}
\newcommand{\githublink}{\href{https://github.com/biographs-ai/grandjury}{GitHub}}
\newcommand{\pypilink}{\href{https://pypi.org/project/grandjury}{\texttt{grandjury}}}

\title{GrandJury: A Collaborative Machine Learning Model Evaluation Protocol for Dynamic Quality Rubrics}
\author{Arthur Cho \\ \texttt{hello@memiorji.com} \\ Memoirji LLC}
\date{\today}

\begin{document}
\maketitle

\begin{abstract}
Generative Machine Learning models have become central to modern systems, powering applications in creative writing, summarization, multi-hop reasoning, and context-aware dialogue~\cite{brown2020language}. These models underpin large-scale AI assistants, workflow automation, and autonomous decision-making~\cite{openai2023gpt4,touvron2023llama}. In such domains, ``acceptable response'' is rarely absolute or static, but plural and highly context-dependent~\cite{dumitrache2018crowdtruth}. Yet standard evaluation regimes still rely on static, benchmark-style tests, incentivizing optimization toward leaderboard scores rather than alignment with dynamic user needs or evolving realities~\cite{ethayarajh2020utility}. \textbf{GrandJury} introduces a formal evaluation protocol—combining time-decayed aggregation, complete traceability, with the support of dynamic, transparent task rubric attribution, and multi-rater human judgment. Together, these elements enable pluralistic, accountable evaluation that captures evolving consensus and surfaces disagreement. We provide (i) an open-source implementation (\texttt{grandjury} PyPI) and (ii) a public collection of Large Language Model (LLM) inference outputs to illustrate the need and method. GrandJury provides a new paradigm for AI practitioners when evaluating machine learning outputs \emph{without absolute ground truth}. Patent pending: US Provisional 63/825,484, filed 17 June 2025.
\end{abstract}

\section{Introduction}

The rapid proliferation of generative AI systems has transformed how information is produced, mediated, and consumed across domains—ranging from creative writing and conversational assistance to automated summarization, analysis, and complex reasoning tasks~\cite{brown2020language,thoppilan2022lamda}. These models now serve as the backbone for agentic processes in digital assistants, enterprise workflows, education, and research~\cite{wornow2024automating}. In such contexts, outputs are not merely ``right'' or ``wrong,'' but reflect a spectrum of valid perspectives, goals, and interpretations~\cite{basile2021disagreement,uma2021learning}.

Yet, the standard machinery for model evaluation remains wedded to the concept of static ground truth data with standardized evaluation criteria. Evaluation metrics such as BLEU~\cite{papineni2002bleu}, ROUGE~\cite{lin2004rouge}, and MMLU~\cite{hendrycks2020measuring} were designed for tasks with single correct answers and well-defined metrics. BLEU and ROUGE, long used for translation and summarization, tend to reward extractive outputs and n-gram overlap, failing to assess abstraction, nuance, or contextual appropriateness~\cite{reiter2018structured}. Applying them to generative, open-ended, or value-laden outputs creates a profound mismatch between what is measured and what actually matters. 

Optimization toward these methodologies often focuses on automating the metrics calculation process. Notable among them is RAGAS (Retrieval-Augmented Generation Assessment)~\cite{es2024ragas}, a suite of reference-free metrics designed to evaluate Retrieval-augmented generation(RAG) systems across dimensions such as retrieval relevance, answer faithfulness, and context precision without requiring human-annotated ground truth. Similarly, LLM-as-judge approaches~\cite{zheng2024judging,ye2024llm}—including frameworks like Prometheus~\cite{kim2024prometheus}, JudgeLM~\cite{zhu2023judgelm}, and GPT-based evaluation pipelines—automate scoring by using models to evaluate other models. Embedding-based techniques such as cosine similarity~\cite{zhang2020bertscore} is also commonly used for automatic plausibility checks. 

However, these proxy methods remain inadequate: RAGAS metrics lack empirical validation for open-ended tasks and often fail to correlate with human preferences~\cite{fabbri2021summeval}, LLM-as-judge systems exhibit variable bias and sensitivity to prompts~\cite{park2024offsetbias,wataoka2024selfpreference}, and embedding-based scores ignore narrative, pragmatic, or stylistic fidelity~\cite{liu2016dialogue}. Thus, while these attempts aim to bypass the need for manual annotation, they still struggle to align with nuanced human judgments. Worse, as vendors and researchers design their own benchmarks, the definition of ``state of the art'' becomes increasingly malleable and self-referential~\cite{lipton2018troubling,fodor2025line}.

\section{Evaluation in the Absence of Certainty}
\subsection{The Limits of Metric-Based Evaluation and the Birth of Vibe Evaluation}
Evaluating generative AI fundamentally differs from closed, deterministic tasks. 
Open-ended outputs introduce further tension: inter-annotator agreement is often low ($\kappa < 0.3$)~\cite{elangovan2024inter,kenyon2018sentiment}, even with multiple raters, because no shared evaluation contract guides judgment consistency. In practice, without clearly defined rubrics, different annotators apply divergent criteria, leading to inconsistency. In such ambiguous domains, evaluation often degenerates into \emph{vibe evaluation}: viral anecdotes or screenshot-based claims that lack reproducibility, structure, or transparency, enabling ``score theater'' and hype to dominate without accountability~\cite{lipton2018troubling}.

\subsection{Attempts at Automated Proxy Evaluation and Their Limitations}

To address evaluation scale challenges, several automated proxies have been proposed:

\begin{itemize}[leftmargin=*]
\item \textbf{RAGAS:} Reference-free evaluation metrics for RAG systems—covering retrieval quality, faithfulness, and context relevance. These metrics are introduced without explicit ground-truth references, but are validated and optimized against human-labeled benchmarks, embedding ground truth indirectly in their design~\cite{es2024ragas}.

\item \textbf{LLM-as-Judge frameworks} rely on models (e.g., GPT-4 or Claude) to rate outputs. While attractive for scalability, studies reveal significant issues: evaluator bias, prompt sensitivity, position bias, and instability—with performance diverging from human agreement in nuanced tasks~\cite{park2024offsetbias,wataoka2024selfpreference}.

\item \textbf{Embedding-based metrics} (e.g., cosine similarity) offer fast, reference-free evaluations but fail to capture deeper semantic, pragmatic, or stylistic coherence, often misaligning with human judgments~\cite{fabbri2021summeval,liu2016dialogue}.
\end{itemize}

Although these methods reduce annotation overhead, they all depend—explicitly or implicitly—on ground-truth signals and still fall short of reliable human-aligned evaluation in open-ended domains.

\section{GrandJury: A Protocol for Structured Judgment}

We introduce GrandJury, a governance protocol designed for pluralistic, transparent, and accountable evaluation in domains lacking definitive ground truth.

\subsection{Consent \& Voter Prompt}

Each task begins with a "voter prompt": a publicly shared rubric specifying what to evaluate, how, and why. Inspired by annotation frameworks like CrowdTruth~\cite{dumitrache2018crowdtruth}—which model annotator disagreement as signal, not noise—this structured rubric aligns diverse judgments under a shared framework while retaining interpretive differences.

\subsection{Time-Decayed Aggregation}

Instead of static averages, GrandJury applies exponential time decay:
\begin{equation}
S_t = \alpha S_{t-1} + (1 - \alpha)\langle v_t \rangle
\end{equation}

\paragraph{Notation.}
\begin{itemize}[leftmargin=*]
  \item $t$ — batch index; $S_{t-1}$ — cumulative score before batch $t$.
  \item $v_{t,i}\!\in\![0,1]$ — vote of juror $i$ in batch $t$ ($1=\text{accept}$; $0=\text{reject}$).
  \item $r_i>0$ — optional reputation weight (defaults to $1$).
  \item $\displaystyle \langle v_t\rangle=\frac{\sum_i r_i v_{t,i}}{\sum_i r_i}$ — reputation-weighted mean.
  \item $\Delta t$ — elapsed time since batch $t{-}1$ (configurable units); $\lambda$ — decay constant (time$^{-1}$).
  \item $\alpha=e^{-\lambda\Delta t}$ — decay factor.
\end{itemize}

\paragraph{Worked example.}
Let $S_{t-1}=0.72$, $\Delta t=3$ days, $\lambda=0.1$, so $\alpha=e^{-0.3}\approx0.741$.
Three jurors vote $\{0.90,0.80,0.60\}$.
Then $\langle v_t\rangle=0.766$ and
\[
S_t=0.741\cdot0.72\;+\;0.259\cdot0.766\;=\;0.733.
\]

\paragraph{Time unit flexibility.}
While this example uses days for illustration, the implementation supports arbitrary time units (seconds, hours, days) with corresponding $\lambda$ values, enabling fine-grained temporal control based on evaluation context.

\paragraph{Parameter selection.}
The decay constant $\lambda$ determines responsiveness: larger values make scores more sensitive to recent votes. Typical values range from $\lambda = 0.01$ time$^{-1}$ (high responsiveness) to $\lambda = 0.001$ time$^{-1}$ (stable consensus), depending on temporal granularity and evaluation requirements.

This echoes ``time-aware'' methodologies in recommendation and human-feedback systems, ensuring recent votes have more influence while preserving historical context.

\paragraph{Batch vs. streaming processing.}
While the formula describes discrete "batches" $t$, the implementation processes votes in real-time: each new evaluation call constitutes a micro-batch, allowing continuous score updates as votes arrive. This architecture supports both individual scoring and population-level analytics through structured vote data containing \texttt{inference\_id}, \texttt{vote}, \texttt{voter\_id}, \texttt{vote\_time}, and \texttt{voter\_prompt\_id}.

\subsection{Freshness Metric}

To maintain clarity on score recency, GrandJury tracks freshness $F_t = 1 - \alpha$, indicating how much weight recent evaluations contribute—borrowing concepts from continual benchmarking and adaptive scoring protocols.

\paragraph{Interpretation.}
Freshness quantifies the influence of \emph{new} evidence:
\[
F_t \;=\;1-\alpha
      \;=\;1-e^{-\lambda\Delta t}.
\]
In the example above, $F_t=0.259$, meaning $25.9\%$ of the current score derives from the latest batch and $74.1\%$ from historical consensus. The resulting score (0.733) demonstrates how the protocol balances stability with responsiveness. Smaller $\Delta t$ or larger $\lambda$ makes the score more responsive to recent evidence.

\subsection{Ambiguity Flagging}

High evaluator variance (e.g., $\sigma^2 > 0.5$) triggers an ambiguity flag, signaling outputs that lack consensus. This transforms disagreement from a problem into an analytical signal, essential for transparent evaluation.

\paragraph{Notation and rule.}
Let $\sigma_t^{2} = \operatorname{Var}(v_{t,*})$.
GrandJury flags an item as \textit{ambiguous} if
\[
\sigma_t^{2} > \sigma_{\text{crit}}^{2},
\qquad \text{e.g. } \sigma_{\text{crit}}^{2}=0.05.
\]

\paragraph{Worked example.}
Using the same three votes $\{0.90,0.80,0.60\}$:
\[
\sigma_t^{2}=\tfrac13\!\sum_i(v_{t,i}-0.766)^2=0.016<0.05,
\]
so \emph{no} flag is raised.  
With divergent votes $\{1,0,0\}$ we get $\sigma_t^{2}=0.111>0.05$; the item is flagged for curator review or an additional voting round.

\subsection{Traceability of Votes}

Every vote is timestamped (ISO 8601), linked to a pseudonymous Juror ID, and associated with its voter prompt context. This mirrors best practices in human evaluation datasheets and crowdsourced auditing, enabling full audit trails and enabling meta-analysis.

\subsection{Implementation Architecture}

The time-decay algorithm is implemented server-side, with the Python client providing data formatting and API access. This architecture ensures:
\begin{itemize}[leftmargin=*]
\item Consistent scoring across all users and evaluation sessions
\item Real-time processing of individual votes as micro-batches
\item Scalable aggregation for large evaluation campaigns
\item Centralized parameter management ($\lambda$, $\sigma_{\text{crit}}^2$) with domain-specific tuning
\end{itemize}

The separation between client-side data handling and server-side computation enables both individual researchers and large-scale evaluation initiatives to leverage the same algorithmic infrastructure while maintaining flexibility in data sources and evaluation workflows.

\section{Demonstration: Public Inference Dataset and Evaluation Toolkit}

To demonstrate GrandJury in practice, we release a comprehensive inference collection comprising 149 joke responses from 31 distinct models across 7 major AI platforms. This dataset serves as both a demonstration of the protocol's applicability and a resource for reproducible evaluation research. The dataset is hosted on \datasetlink.

\subsection{Model Selection Methodology}

Model selection followed a systematic approach to ensure representation of high-performing models in creative tasks. We consulted the LMSYS Chatbot Arena leaderboard~\cite{lmsys2024arena}, specifically filtering for models that demonstrated excellence in creative writing tasks. From this curated list of top-performing creative models, we selected those available across major cloud AI platforms (Azure AI, Anthropic, Bedrock, Gemini, Hugging Face, Mistral, and NVIDIA) to ensure accessibility and reproducibility for the research community.

This methodology resulted in 31 distinct models spanning multiple model families and sizes, providing a representative sample of state-of-the-art generative capabilities in creative domains. All models were queried with the identical prompt ``tell me a joke'' using temperature=1.0 for consistent experimental conditions. The final collection comprises 149 responses collected within a controlled 6-minute window, ensuring temporal consistency across platforms.

\subsection{Dataset Characteristics and Design}

The inference collection contains comprehensive metadata for each response:
\begin{itemize}[leftmargin=*]
\item Model identifier and provider platform
\item Complete inference input and generated output
\item ISO-8601 timestamps for temporal analysis
\item Platform-specific model configurations and parameters
\end{itemize}

Table~\ref{tab:dataset_sample} presents a representative sample from the inference collection, showcasing the diversity of models, platforms, and response styles captured in the dataset. Each response represents a distinct model's interpretation of the identical prompt ``tell me a joke'' with temperature=1.0, illustrating the natural variation in generative outputs that motivates the need for pluralistic evaluation.

\begin{table}[ht]
\centering
\caption{Representative Sample from GrandJury Inference Collection}
\label{tab:dataset_sample}
\begin{tabular}{p{2.2cm}p{3.2cm}p{2.8cm}p{6cm}}
\toprule
\textbf{Platform} & \textbf{Model} & \textbf{Timestamp} & \textbf{Response} \\
\midrule
Azure AI & gpt-4.1 & 2025-08-01 22:57:27 & Why did the scarecrow win an award? Because he was outstanding in his field! \\
\addlinespace
Anthropic & claude-opus-4-20250514 & 2025-08-01 22:57:27 & Why don't scientists trust atoms? Because they make up everything! \\
\addlinespace
Mistral & mistral-medium-2505 & 2025-08-01 22:57:26 & Sure! Here's a classic for you: \textbf{Why don't skeletons fight each other?} \textit{Because they don't have the guts!} \\
\addlinespace
Vertex AI & gemini-2.5-pro & 2025-08-01 22:57:37 & Why don't scientists trust atoms? Because they make up everything \\
\addlinespace
NVIDIA NIM & qwen/qwen3-235b-a22b & 2025-08-01 22:57:32 & Sure! Here's a light-hearted one for you: Why did the tomato turn red? Because it saw the salad dressing! \\
\addlinespace
Bedrock & nova-pro-v1:0 & 2025-08-01 22:57:27 & Sure, here's a classic joke for you: Why did the scarecrow win an award? Because he was outstanding in his field! \\
\addlinespace
Hugging Face & Qwen/Qwen3-30B-A3B & 2025-08-01 22:57:29 & Why did the scarecrow win an award? \textbf{Because he was outstanding in his field!} \\
\bottomrule
\end{tabular}
\end{table}

By design, there is no single ``correct'' output for the humor generation task, and the dataset is released without reference answers or gold standards. This structure deliberately underscores the limitations of similarity-based metrics (e.g., cosine embedding similarity), which have repeatedly been shown to misalign with human judgments in subjective domains such as humor, creativity, and open-ended reasoning.

The dataset is publicly available via Hugging Face to foster reproducible, community-driven evaluation. The GrandJury Inference Collection—\emph{Jokes on Us}\footnote{\href{https://huggingface.co/datasets/biographs/jokes_on_us}{biographs/jokes\_on\_us} (v1.0, CC-BY-4.0, accessed 4 Aug 2025).} is released under the \textbf{CC-BY-4.0} licence and is archived on Hugging Face for reproducibility~\cite{biographs_2025_jokes}.

\subsection{Two-Phase Evaluation Workflow}

GrandJury operates through a clear separation of concerns between data collection and vote analysis:

\textbf{Phase 1: Human Evaluation Collection}
\begin{enumerate}[leftmargin=*]
\item Download the inference dataset from Hugging Face
\item Define voter prompts (evaluation rubrics) appropriate for humor assessment
\item Conduct human evaluation sessions to collect votes on model outputs
\item Store votes with required metadata (voter ID, timestamps, inference ID)
\end{enumerate}

\textbf{Phase 2: Vote Analysis with grandjury Package}
\begin{enumerate}[leftmargin=*]
\item Load collected vote data using the grandjury Python client
\item Apply time-decay algorithms and consensus analysis
\item Generate evaluation reports with traceability and freshness metrics
\end{enumerate}

This separation ensures that evaluation criteria remain flexible and community-driven, rather than embedded in automated metrics. The inference dataset serves as raw material for human evaluation, while the grandjury package provides the analytical infrastructure for processing the resulting human judgments.

Researchers and practitioners are invited to explore the dataset and apply the GrandJury protocol to this inference collection: by defining voter prompts, collecting plural judgments, and aggregating scores using the time-decay method, the community can surface disagreements, map consensus shifts, and transparently assess generative models beyond static benchmarks. 

We provide a reference implementation of the GrandJury protocol through the open-source \pypilink{} Python package~\cite{grandjury_2025_pypi} (available on PyPI and GitHub). This package is designed for pipeline-ready use, enabling researchers and practitioners to adopt the protocol for vote analysis and time-decay scoring.

\subsection{Package Installation}

Install via pip:
\begin{verbatim}
pip install grandjury
\end{verbatim}

Optional performance dependencies for enhanced data format support:
\begin{verbatim}
pip install grandjury[performance]  # Includes msgspec, pyarrow, polars
\end{verbatim}

Source code and documentation are available at \githublink.

The package structure follows established design patterns in scientific Python libraries (e.g., Scikit-learn) to ensure familiarity and ease of adoption.

\subsection{Vote Analysis API}

The package operates on vote collections gathered during Phase 1 evaluation. Users can analyze vote data across multiple formats:

\begin{verbatim}
from grandjury import GrandJuryClient
import pandas as pd

# Initialize client (no authentication needed for analysis)
client = GrandJuryClient()

# Load vote data from your evaluation sessions
vote_data = pd.read_csv("collected_votes.csv")

# Analyze vote patterns and temporal distributions  
histogram = client.vote_histogram(vote_data)
completeness = client.vote_completeness(vote_data, voter_list)
confidence = client.population_confidence(vote_data, voter_list)
distribution = client.votes_distribution(vote_data)

# Apply time-decay scoring (requires API key for server access)
client = GrandJuryClient(api_key="your-api-key")
result = client.evaluate_model(
    previous_score=0.0,
    votes=[0.8, 0.6, 0.9],  # From your collected evaluations
    reputations=[1.0, 1.0, 1.0]
)
print(f"Time-decayed score: {result['score']:.4f}")
\end{verbatim}

The package supports multiple data formats (pandas DataFrames, Polars, CSV, Parquet, JSON) to integrate seamlessly with existing evaluation workflows.

\subsection{Ecosystem Integration \& Extensibility}

\texttt{grandjury} is intentionally designed as a reference implementation, with minimal dependencies and a modular API to facilitate integration into broader tooling—such as dashboards, analytics pipelines, or UI plugins.

Researchers can:
\begin{enumerate}[leftmargin=*]
\item Apply the package to community or proprietary datasets,
\item Extend pipelines in custom repositories,
\item Leverage time-decay and freshness functions in other evaluation systems.
\end{enumerate}

This positions \texttt{grandjury} not only as a standalone tool but as foundational infrastructure for structured, pluralistic evaluation.

\section{Conclusion and Community Engagement}

GrandJury is not a benchmark, metric, or product—it is a protocol and community infrastructure designed for pluralistic, transparent, and renewable evaluation of generative AI where ground truth doesn't apply. This final section crystallizes the purpose, situates it within current governance dynamics, and issues a strong, actionable invitation to engage.

\subsection{Contributions and Objectives}

GrandJury's core aim is to enable structured, pluralistic evaluation, not to declare winners or embed opaque rankings. Its emphasis on plurality over consensus, transparency over opacity, and renewal over stagnation aligns with emerging AI governance frameworks advocating for independent evaluation and human-centric oversight mechanisms.

\subsection{Protocol Scope, Community Adoption, and Future Directions}

GrandJury serves as a governance protocol that defines who votes, under which voter prompt, when, and how those votes are aggregated, while functioning as a system to surface plural judgment, document metadata, and track consensus changes over time. Importantly, it is not a static scoring metric, commercial SaaS product, or prescriptive policy—rather, it provides tooling and structure for evaluation without imposing directives. This positioning reflects our commitment to enabling structured, pluralistic evaluation without declaring winners or embedding opaque rankings.

We invite broad community engagement across multiple stakeholder groups: researchers and practitioners to apply GrandJury on their models, domains, and datasets; open-source developers to build UIs, dashboards, and workflow integrations on top of the protocol; governance bodies, standards groups, and auditors to pilot GrandJury in transparent, real-world evaluation or compliance use cases; and human-AI ethics and policy researchers to study and refine the protocol's transparency, accountability, and consensus tracking dynamics. This collaborative approach ensures that evaluation criteria remain flexible and community-driven rather than embedded in automated metrics.

GrandJury's design aligns with current institutional efforts toward responsible AI governance. The protocol echoes third-party and audit-based evaluation systems, such as the ISO/IEC 42001 AI Management System standard promoted by the British Standards Institution for independent auditing and certification of AI governance processes~\cite{iso42001}, while supporting responsible AI initiatives that prioritize transparency, human oversight, and auditable evaluation~\cite{nist2023aiframework,oecd2023accountability}. Its emphasis on plurality over consensus, transparency over opacity, and renewal over stagnation resonates with regulatory movements like the EU AI Act, which mandate human oversight and independent conformity assessment~\cite{eu2025aiact}. By anchoring evaluation in plural human judgments—logged, auditable, and time-aware—GrandJury offers infrastructure that embraces uncertainty, decentralizes authority, and transforms disagreement into accountability.

Looking forward, we urge the community to download the Inference Collection, experiment with voter prompts, and apply the \texttt{grandjury} package in practice. We encourage sharing datasets, voter-prompt variants, and insights publicly—via GitHub, forums, or publications—while engaging in protocol dialogues that refine metadata logging, decay parameters, and ambiguity thresholds. The ultimate goal is integrating GrandJury within organizational audit workflows and community evaluation standards, positioning it not only as a standalone tool but as foundational infrastructure for structured, pluralistic evaluation across diverse domains where ground truth remains elusive.

By anchoring evaluation in plural human judgments—logged, auditable, and time-aware—GrandJury offers a new foundation for evaluating generative systems: one that embraces uncertainty, decentralizes authority, and transforms disagreement into accountability.


\end{document}